\documentclass[review]{elsarticle}

\usepackage{lineno,hyperref}
\usepackage{graphicx}
\usepackage{amsmath,amssymb} 
\usepackage{enumerate}
\usepackage{wrapfig}
\usepackage{multicol}  
\usepackage{multirow}
\usepackage{subfigure}
\usepackage{caption} 
\usepackage{booktabs} 
\usepackage{makecell}
\usepackage{url}  
\usepackage{bbm} 
\usepackage{indentfirst}
\usepackage[ruled,linesnumbered]{algorithm2e}

\newcommand{\ie}{\textit{i}.\textit{e}. }

\modulolinenumbers[5]

\journal{Journal of \LaTeX\ Templates}

\begin{document}

\begin{frontmatter}

\title{BPFNet: A Unified Framework for Bimodal Palmprint Alignment and Fusion}

\author[mymainaddress1]{Zhaoqun~Li}
\ead{zhaoqunli@link.cuhk.edu.cn}
 
\author[mymainaddress6]{Xu~Liang} 
\ead{xuliangcs@gmail.com}

\author[mymainaddress2]{Dandan~Fan}
\ead{fanfan\_abu@163.com}

\author[mymainaddress6]{Jinxing~Li}
\ead{lijinxing158@gmail.com}  

\author[mymainaddress1,mymainaddress2]{David~Zhang\corref{mycorrespondingauthor}}
\cortext[mycorrespondingauthor]{Corresponding~author}
\ead{davidzhang@cuhk.edu.cn}

\address[mymainaddress1]{The Chinese University of Hong Kong, Shenzhen}

\address[mymainaddress2]{Shenzhen Institute of Artificial Intelligence and Robotics for Society}
\address[mymainaddress6]{Harbin Institute of Technology, Shenzhen}

\begin{abstract}
	Bimodal palmprint recognition leverages palmprint and palm vein images simultaneously,
	which achieves high accuracy by multi-model information fusion and has strong anti-falsification property.
	In the recognition pipeline, 
	the detection of palm and the alignment of region-of-interest (ROI) are two crucial steps for accurate matching.
	Most existing methods localize palm ROI by keypoint detection algorithms,
	however the intrinsic difficulties of keypoint detection tasks make the results unsatisfactory. 
	Besides, the ROI alignment and fusion algorithms at image-level are not fully investigaged.
	To bridge the gap, 
	in this paper, 
	we propose Bimodal Palmprint Fusion Network (BPFNet) which focuses on ROI localization, alignment and bimodal image fusion.
	BPFNet is an end-to-end framework containing two subnets: 
	The detection network directly regresses the palmprint ROIs based on bounding box prediction and conducts alignment by translation estimation.
	In the downstream,
	the bimodal fusion network implements bimodal ROI image fusion leveraging a novel proposed cross-modal selection scheme. 
	To show the effectiveness of BPFNet,
	we carry out experiments on the large-scale touchless palmprint datasets CUHKSZ-v1 and TongJi
	and the proposed method achieves state-of-the-art performances.
\end{abstract}

\end{frontmatter}

\section{Introduction}
\begin{table*}[t]
\begin{center}
    \scalebox{1}[1]{
    \begin{tabular}{|c|c|c|c|c|c|c|}
    \hline
    Dataset										& Year	& Hands	& Images	& Environment	& \#Keypoint  \\
    \hline
    \hline
    CASIA\cite{CASIA}							& 2005	& 624 	& 5,502 	& Constrained	& -				\\
    \hline
    IITD-v1\cite{IITD}							& 2006	& 460 	& 3,290 	& Constrained	& -				\\
    \hline
    TongJi\cite{zhang2017towards}				& 2017	& 600 	& 12,000 	& Constrained	& -				\\
    \hline
    NTU-CP-v1\cite{matkowski2019palmprint} 		& 2019  & 655 	& 2,478 	& Unonstrained	& 9				\\
    \hline
    MPD-v2\cite{zhang2020towards}				& 2020	& 400 	& 16,000 	& Unonstrained 	& 4				\\
    \hline
    XJTU-UP\cite{shao2020towards}          		& 2020  & 200 	& $>$20,000 & Unonstrained	& 14			\\
    \hline
    CUHKSZ-v1\cite{li20213d}					& 2021	& 2334 	& 28,008 	& Unonstrained	& 4				\\
    \hline
\end{tabular}}
\end{center}
\caption{Touchless palmprint datasets.}
\label{tab_dataset}
\end{table*}

Biometric aims to identify individual by his/her intrinsic attributes,
including iris \cite{nalla2016toward,nguyen2017long}, 
face \cite{Schroff2015facenet,Opitz2016grid,deng2019arcface}, 
fingerprint \cite{Cappelli2010Minutia,maltoni2009handbook}, and palmprint. 
As a representative technology in biometric, 
touchless palmprint recognition has drawn great attention to researchers recently due to its potential applications on person identification.
A pioneer work of palmprint recognition is PalmCode \cite{Zhang2003Online} that uses orientation information as features for matching. 
After that, 
more and more coding based methods \cite{Kong2004Competitive,jia2017palmprint} emerge in this field.
With the development of machine learning algorithms, 
researchers bend their effort for extracting high discriminative palmprint descriptor via leveraging local region information \cite{luo2016local},
collaborative representation \cite{zhang2017towards}, 
SIFT operator \cite{charfi2016local}, binary representation \cite{fei2019learning} and so on.
Recently, 
convolutional neural network (CNN) has achieved tremendous success in palmprint related tasks such as palmprint alignment \cite{matkowski2019palmprint},
hyperspectral palmprint verification \cite{zhao2019joint} and palmprint ROI feature extraction \cite{Genovese2019palmnet}.
Inspired by powerful metric learning techniques \cite{hoffer2015deep,deng2019arcface} in face recognition, 
\cite{liu2020contactless,zhu2020boosting} employ well-designed loss functions to enhance intra-class and inter-class distance distribution.
 
Among diverse application schemes,
bimodal palmprint recognition takes advantage of palmprint and palm vein images simultaneously and achieves better performance.
Compared to person identification using single palmprint,
palmprint recognition with dual-camera could improve recognition accuracy by multi-model information fusion and has high anti-falsification property.
In the bimodal palmprint recognition pipeline,
palm detection and ROI extraction are essential prerequisites for feature extraction and have a large influence on the final performance.
In \cite{matkowski2019palmprint}, 
an end-to-end VGG \cite{simonyan2014very} based framework is proposed for joint palmprint alignment and identification.
Nevertheless,
the model size is huge which is not suitable for mobile or other real-time applications.
\cite{zhang2020towards} adopts a YOLOv3 based detector for palm detection while its keypoint detection strategy is not always stable and the model is not designed for multi-modal image fusion.
Most existing works fulfill ROI extraction by keypoint detection,
however the task is not robust and generally more difficult compared to bounding box regression.

The palmprint is usually an RGB image and the palm vein is an infra-red (IR) image.
In palmprint recognition,
the intrinsic disparity between RGB and IR images makes the ROI alignment crucial for accurate matching.
However, 
most public bimodal datasets do not contain adequate camera information for ROI alignment,
thus the alignment is restricted at image-level. 
Moreover,
bimodal palmprint feature fusion is an unavoidable module which is responsible for exploit vital information between two images.
A well-designed fusion scheme is desired to boost the recognition performance.

To address the above concerns,
we aim to design a unified framework for efficient bimodal palmprint recognition.
In this paper,
we propose our BPFNet which conducts ROI localization,
ROI alignment and bimodal feature fusion tasks with an end-to-end fashion.
In BPFNet, 
the detection network (DNet) directly regresses the rotated bounding box by point view and also predicts the image disparity,
making it compatible with other annotation systems.
After extracting the ROIs in a differentiable manner,
the fusion network (FNet) refines the palmprint and palm vein ROI features, obtaining the final descriptor
by a novel cross-modal selection mechanism.
Finally, 
experimental results and ablation studies on two datasets demonstrate the superiority of our method.
To summarize, the contribution of our paper is three-fold:
\begin{enumerate}
	\item We propose a novel end-to-end deep learning framework fulfilling palm detection, ROI alignment and bimodal feature fusion simultaneously,
	which can generate a high discriminative palmprint descriptor. 
	\item A novel ROI localization scheme is applied, 
	which is also compatible with other datasets,
	achieving $90.65\%$ and $\%$ IoU on CUHKSZ and TongJi datasets respectively.  
	\item We design a novel cross-modal selection module for bimodal image fusion,
	where the fusion is dominated by the palmprint feature and the selection is based on the correlation between image features.
\end{enumerate}

The rest of this paper is organized as follows. 
The related works describe briefly describe selected palmprint feature extraction and matching methods.
Some popular touchless palmprint databases and corresponding ROI extraction algorithms ares also introduced. 
The principle of our ROI extraction and alignment algorithm is depicts in Section \ref{sec_pre}.
Our proposed framework BPFNet as well as its components and inference are analyzed in Section \ref{sec_method}, 
followed by the experimental analysis in Section \ref{sec_experiment}. 
This paper is finally concluded in Section \ref{sec_conclusion}.

\section{Related Works}
\label{sec_rw}
In this section, 
we first introduce some touchless palmprint benchmarks and related ROI localization algorithms.
Next, we review some palmprint recognition methods, 
where we will focus on recent progress on machine learning and deep learning approaches.

\subsection{Touchless Palmprint Benchmarks and Corresponding ROI Extraction Methods.}
\label{sec_rw_dataset}
With the development of touchless palmprint recognition technology,
different kinds of databases are established by the community.
The earlier touchless palmprint datasets are CASIA and IITD-v1 which are released by the Chinese Academy of Sciences \cite{CASIA}
and the Indian Institute of Technology in Dehli \cite{IITD} respectively.
IITD-v1 contains 3,290 palmprint images from 460 palms and the images are captured in a stable and uniform environment.
The acquisition device of CASIA creates a semi-close environment and capture 5,502 gray-scale palmprint images from 624 hands.
For obtaining high quality images,
the data collection of TongJi \cite{zhang2017towards} (12,000 images) is conducted in a more constrained environment.
The volunteers need to put their hands in a semi-box and the positions of fingers is guided by the device.
The above datasets follow \cite{Zhang2003Online} to locate the region-of-interest (ROI) region, 
which is based on finding landmarks on the extracted hand contour.
To be specific, 
after binarizing the palmprint using an appropriate threshold,
\cite{Zhang2003Online} applies a line scanning method to detect the gap between the index and the middle fingers and the gap between the ring and the little fingers.
Then the ROI is determined by a local coordinate system which is built based on the two finger gaps.
However since the method relies on the binarization of images,
the ROI extraction is not stable for RGB images, especially in complex scenes.

With the development of general object detection \cite{Ren2017faster,He_2017_ICCV,yolov3,zhou2019objects},
more and more researchers turn to leverage learning based methods. 
Recent works \cite{matkowski2019palmprint,zhang2020towards,shao2020towards,li20213d} aim to define the ROI with the aid of manual annotatation of keypoints on the palmprint images.
NTU-CP-v1 \cite{matkowski2019palmprint} contains 2478 images from 655 palms of 328 subjects.
On each image 9 landmarks are labeled for correcting possible elastic palm deformations caused by different hand poses.
To fulfill the alignment task, 
the proposed ROI-LANet employs a Spatial Transformer Network \cite{Jaderberg2015Spatial} with TPS \cite{bookstein1989principal} to regress the landmarks.
The dataset MPD-v2 \cite{zhang2020towards} marks four successive finger-gap-points and the proposed method adopts a YOLOv3 \cite{yolov3} based model to detect the double-finger-gap points and the palm-center point.
The ROI is then extracted by a local coordinate system.
For XJTU-UP \cite{shao2020towards},
there are 14 keypoints on the palm contour, 
including 3 valley points between fingers, 8 points at the bottom of fingers, and 3 points on either side of the palm.
CUHKSZ-v1 \cite{li20213d} has both palmprint RGB images and palm vein IR images,
where only the RGB images are annotated.
The four joints between fingers and palm are marked as keypoints for locating ROI.
There are totally 2334 hand images from 1167 individuals in this dataset.
The details of the touchless datasets are summarized in Table \ref{tab_dataset}.

\subsection{Palmprint Recognition Methods}
\label{sec_rw_method}
According to how the kernel filters are obtained,
the existing methods could be coarsely grouped into two categories: 
conventional methods and CNN based methods.

Coding based methods are the most popular approaches in past decades.
The representative works PalmCode \cite{Zhang2003Online} and Competitive Code \cite{Kong2004Competitive} 
leverage Gabor filters to extract line features on s and conduct per pixel matching.
In the feature extraction process,
the orientation information is encoded efficiently in the feature map.
To overcome the weakness of per pixel matching,
many region based methods that utilize the local statistic information emerge.
LLDP \cite{luo2016local} splits the whole feature map into several blocks and a histogram based distance is calculated in the matching. 
For determining the optimal direction feature,
LDDBP \cite{fei2019localdiscriminant} applies an exponential Gaussian fusion model to generate a local binary descriptor,
which achieves high performance.
CR\_CompCode \cite{zhang2017towards} is a machine learning based method that also considers the samples in the training gallery.
The proposed collaborative palmprint representation has high accuracy in the person verification task and the matching process is fast. 
In \cite{jia2017palmprint}, 
a general framework for direction representation based method is proposed,
where complete and multiple features are ensembled according to the correlation and redundancy among them. 

Recent development of deep learning also brings tremendous success in palmprint related tasks such as palmprint alignment \cite{matkowski2019palmprint},
hyperspectral palmprint verification \cite{zhao2019joint} and palmprint ROI feature extraction \cite{Genovese2019palmnet}.
\cite{liu2020contactless} design a faster-RCNN \cite{Ren2017faster} based architecture to detect palmprint regions and design a adapted triplet loss function to optimize distance distribution.
Motivated by powerful metric learning techniques, 
\cite{zhu2020boosting} propose an adversarial metric learning approach to optimize the distance distribution in the hypersphere embedding space.
In order to realize palmprint recognition in mobile devices,
\cite{zhang2020towards} adopts a YOLOv3 based detector for palm detection and another backbone for ROI extraction and matching,
achieving high accuracy and the inference latency is well controlled.

\section{Preliminaries}
\label{sec_pre}

\begin{figure}[t]
	\centering
	\includegraphics[width=\linewidth]{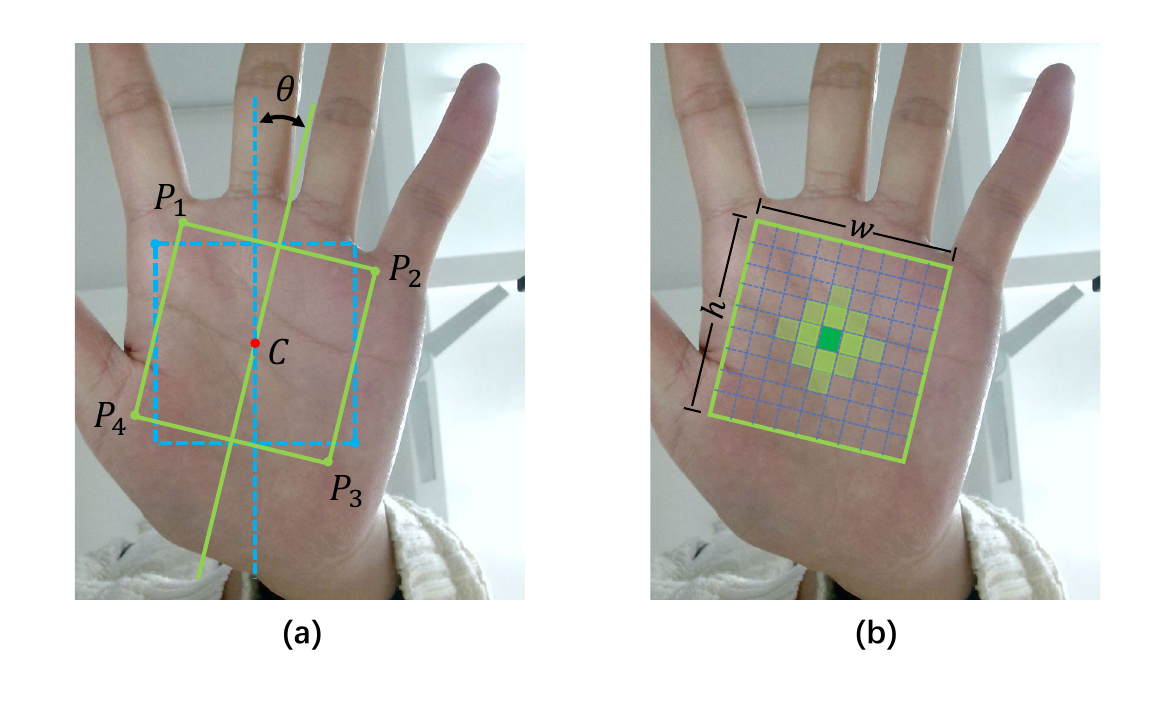}
	\caption{Illustration of ROI localization and ground truth heat map.
	(a) The ground truth ROI (green box) can be regard as a rotated version of blue box, 
	which could be represented as $(x_c,y_c,w,h,\theta)$.
	(b) The ground truth heat map is a gaussian distribution centered on palm center $C$ 
	and the standard deviation is determined by the box size.
	}
	\label{fig_center} 
\end{figure}

\subsection{ROI Localization}
The palmprint ROI is a rotated rectangle (generally a square) which is determined by keypoints on the hand image.
Touchless palmprint datasets may have different annotation systems and the number of keypoints also varies (Table \ref{tab_dataset}).
In order to generalize our detection algorithm,
instead of detecting the key points,
we regress the ROI with a rotated bounding box directly.
Denote the center of ROI bounding box as $C(x_c,y_c)$, 
as shown in Fig. \ref{fig_center}(a),
the ground truth bounding box (green box) can be obtained by rotating a regular bounding box (blue box) around $C$ by angle $\theta$.
Therefore we can represent the bounding box as $(x_c,y_c,w,h,\theta)$,
where $w$ and $h$ denote its width and height.
In our methods, 
we employ the DNet to learn these parameters,
which will be described in Section \ref{sec_detect}. 

\subsection{ROI Alignment}
\begin{figure}[t]
   \centering
   \includegraphics[width=\linewidth]{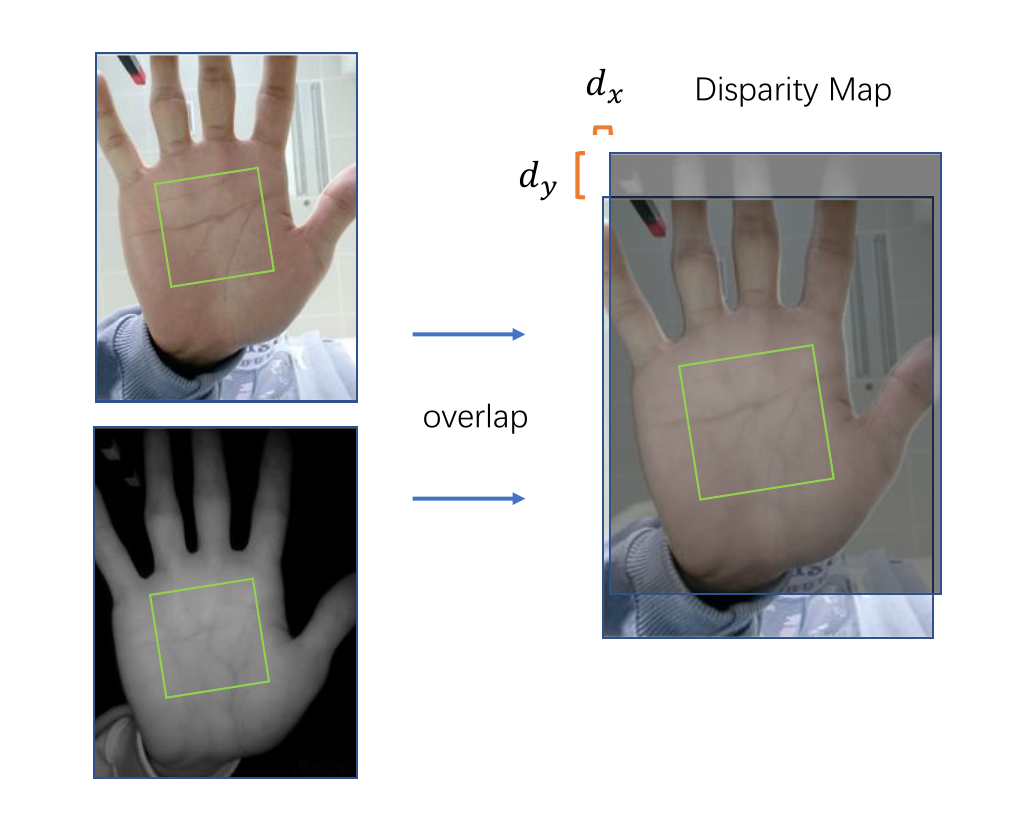}
   \caption{Disparity determination by overlapping images. 
   In CUHKSZ dataset,
   the disparity between images could be formuated as translations.}
   \label{fig_disparity} 
\end{figure}

In CUHKSZ dataset,
only RGB images are annotated while the ground truth ROIs for palm vein images are not available.  
Fortunately, 
since the two cameras are fixed on the same plane in the data acquisition process,
the disparity between the captured images depends only on the height of the hand \cite{Liang2019A}.
Thus we could determine the disparity in image level by overlapping two images,
as shown in Fig. \ref{fig_disparity}.

Assuming the hand is open and flat (as requested in the acquisition process) in the image,
the misalignment between palmprint and palm vein can be formulated as the translation in pixels $(d_x, d_y)$. 
For obtaining accurate information of translation disparity,
we design an automated algorithm which segments the hand in RGB image by skin color,
as described in Algorithm \ref{alg_mask}.
On the other side,
we use OTSU algorithm \cite{otsu1979threshold} to binarize IR image.
After obtaining two hand masks $M_{rgb}$ and $M_{ir}$,
we overlap them and slide one of them in two directions.
The disparity is determined when two masks have maximal intersection area.

\begin{algorithm}[tb] 
	\caption{Palm semgmentation of RGB image}
	\label{alg_mask}
	\SetKwData{Left}{left}
	\SetKwFunction{Erode}{Erode}\SetKwFunction{Dilate}{Dilate}
	\KwIn{Palmprint image $I_{rgb}$}
	\KwOut{Hand mask $M_{rgb}$}
	\BlankLine
	Extract the ROI image $I_{roi}$ from $I_{rgb}$ by annotatation information\;
	Calculate the mean value $\mu = (\mu_r,\mu_g,\mu_b)$ and the standard deviation $\sigma = (\sigma_r,\sigma_g,\sigma_b)$ of each channel in $I_{roi}$\;
	\ForEach{pixel $p$ in $I_{rgb}$} 
	{\eIf{$\mu-3\sigma < p < \mu+3\sigma$}
		{$M_{rgb}=1$\;}
		{$M_{rgb}=0$\;}
	}
	$M_{rgb} \leftarrow$  \Dilate{$M_{rgb}$}\;
	$M_{rgb} \leftarrow$  \Erode{$M_{rgb}$}\;
\end{algorithm}

\begin{figure*}[t]
	\centering
	\includegraphics[width=\textwidth]{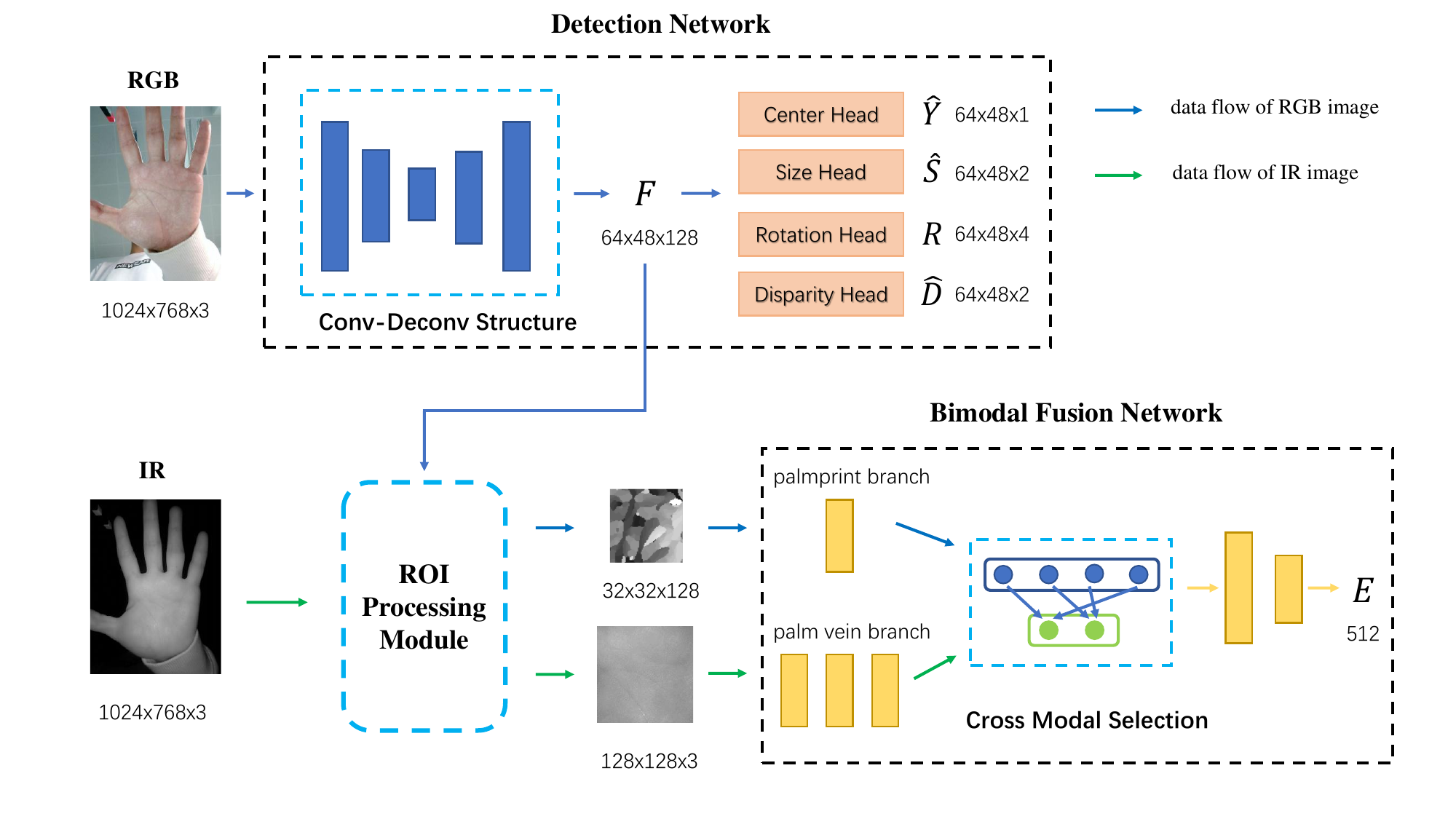}
	\caption{The pipeline of BPFNet. 
	The feature size is tagged with the format $H \times W \times C$.
	DNet generates palmprint feature $F$ using a ResNet backbone and four heads are appended which are designed for ROI prediction and disparity estimation.
	The predicted parameter is then passed to ROI Processing Module for ROI extraction and alignment.
	In the downstream, both palmprint ROI and palm vein ROI are input to FNet.
	The final palmprint descriptor $E$ in FNet is generated by cross-modal selection.}
	\label{fig_pipeline} 
\end{figure*}

\section{Method}
\label{sec_method}

The goal of BPFNet is to realize an accurate palmprint recognition framework with an end-to-end fashion.
The pipeline of our method contains three parts: Detection Network (DNet), 
ROI Processing Module and Bimodal Fusion Network (FNet),
which are shown in Fig. \ref{fig_pipeline}.
In this section, 
we will describe these parts in detail,
where the novel four heads design in ROI localization and the proposed cross-modal selection mechanism in feature fusion are emphatically introduced.

\subsection{Detection Network}
\label{sec_detect}
Most existing works adopt keypoint detection based methods to recover palmprint ROI on the hand image,
however the inherent difficulties of keypoint detection task make the ROI localization algorithm unstable and unreliable,
especially under a complicated environment.
This phenomenon is also observed in \cite{liu2020contactless,zhu2020boosting}.
With various annotation systems across datasets,
it is natural to ask a question:  
\textit{Can we detect the palmprint ROI directly unless the annotation systems?}
And hence the issues in keypoint detection can be mitigated and ROI localization across datasets can be unified in one single framework and 
Following the idea, 
we propose a CenterNet \cite{zhou2019objects} based detector that directly regresses the rotated ROI bounding box.

\textbf{Backbone.}
The design of the backbone network follows \cite{xiao2018simple} which augments ResNet \cite{he2016deep} with deconvolution operators to adapt to the detection task.
In our DNet,
the front part is a ResNet18 network whose output stride is 64,
following two deconvolution layers with batch normalization \cite{ioffe2015batch} and ReLU activation. 
The output stride of the deconvolution layers is 2, so the final output stride is 16.
Denote the backbone feature as $F \in \mathbb{R}^{H\times W\times C_B}$,
where $H, W$ are the height and width of the feature map.
For the sake of memory usage,
in BPFNet the backbone feature $F$ is reused for bimodal fusion and palmprint recognition. 

\textbf{Four Heads Design.}
There are three heads in \cite{zhou2019objects} which are responsible for center prediction,
box size prediction and offset prediction respectively.
In this paper, 
the offset head is dropped since we find it has no influence on performance.
In our detection network,
we have four heads on the top of the backbone feature $F$, 
namely \textit{Center Head, Size Head, Rotation Head,} and \textit{Disparity Head},
as shown in Fig. \ref{fig_pipeline}. 
Each head contains a Conv-ReLU-Conv subnet and generates a heat map whose shape is the same as $F$.
The first three heads are responsible for predicting the rotated bounding box and the last is to estimate the disparity between palmprint image and palm vein image.

For the palm center $C$, 
we compute a low-resolution equivalent $\tilde{C} = \lfloor \frac{C}{16} \rfloor$.
Each pixel on the feature maps contains predicted values and the pixel on the palm center should have the most information about the palm.
Let $\hat{Y} \in \mathbb{R}^{H\times W}$ be the output heat map of the Center Head,
of which the palm center $(\tilde{x}_c, \tilde{y}_c)$ is expected to be 1 while other position should be 0.
The ground truth heat map $Y \in \mathbb{R}^{H\times W}$ is further splat by a Gaussian kernel
$Y_{xy} = \exp(-\frac{(x-\tilde{x}_c)^2+(y-\tilde{y}_c)^2}{2\sigma^2})$,
where $\sigma$ is an adaptive standard deviation depending on the box size \cite{zhou2019objects}.
The formation of $Y$ is illustrated in Fig. \ref{fig_center}(b) .
We use pixel-wise logistic regression with focal loss \cite{lin2017focal} as the supervision signal for the palm center prediction:

\begin{equation}
	L_c = -\sum_{xy}\left\{
	\begin{aligned}
	&(1-\hat{Y}_{xy})^\alpha \log(\hat{Y}_{xy}) & {\rm if} \ Y_{xy}=1 \\
	&(1-\hat{Y}_{xy})^\beta (\hat{Y}_{xy})^\alpha \log(1-\hat{Y}_{xy}) &  \rm{otherwise}
	\end{aligned}
	\right.
\end{equation}
where $\alpha, \beta$ are hyperparamters to adjust the loss curve.
In this work, 
we follow the original paper \cite{lin2017focal} and fix their values $\alpha=2, \beta=4$.

As to the Size Head and Disparity Head, 
we regress the target ground truth values $S = (w,h)$ and $D = (d_x,d_y)$ on the palm center $(\tilde{x}_c, \tilde{y}_c)$.
The regression is supervised by L1 loss functions:
\begin{equation}
\begin{aligned}
	L_s &= |\hat{S}_{\tilde{x}_c\tilde{y}_c}-S| \\
	L_d &= |\hat{D}_{\tilde{x}_c\tilde{y}_c}-D|
\end{aligned}
\end{equation}
where $\hat{S} \in \mathbb{R}^{H\times W\times 2}$ and $\hat{D} \in \mathbb{R}^{H\times W\times 2}$
denote the size prediction and disparity prediction feature maps.

Besides the above heads, 
we add a Rotation Head for inclination prediction.
Since the direct regression for $\theta$ is relatively hard \cite{Mousavian_2017_CVPR},
we take a strategy that first judges whether the orientation is positive and then selects the output angular value. 
To be specific,
each pixel in the feature map $R$ outputs 4 scalars,
where the first two logits are used to conduct orientation classification and the rest scalars are corresponding angular values.
Suppose $R_{\tilde{x}_c\tilde{y}_c}=(l_1, l_2, \theta_1, \theta_2)$,
the classification is trained with softmax loss and the angular values are trained with L1 loss:

\begin{equation}
\begin{aligned}
	L_r = \sum_{i=1}^2 softmax(l_i, c_i) + c_i | \theta_i - \theta|
\end{aligned}
\end{equation}
where $c_i = \mathbbm{1}(\theta>0)$ is the sign indicator.

\subsection{ROI Processing Module}
The four heads enable us to recover the rotated bounding box and extract the ROI on backbone feature $F$ and palm vein image.

In the ROI recovery, 
we pick the pixel in $\hat{Y}$ with maximum value as palm center, denoted as $(\hat{x}, \hat{y})$.
Then the predicted bounding box is simply $(\hat{x}, \hat{y}, \hat{w}, \hat{h}, \hat{\theta})$.
After obtaining the bounding box,
we need to crop the ROIs of palmprint and palm vein images for further bimodal fusion and recognition.
To limit the computational burden,
the ROI extraction of palmprint is operated on the backbone feature $F$. 
In our implementation,
$F$ is first rotated by angle $\theta$ which is realized by applying affine transformation matrix $T$:
$$ T = 
\begin{bmatrix} 
	\cos\theta	& \sin\theta	& (1-\cos\theta)\cdot \tilde{x}_c - \sin\theta\cdot \tilde{y}_c \\ 
	-\sin\theta & \cos\theta 	& \sin\theta\cdot \tilde{x}_c - (1-\cos\theta)\cdot \tilde{y}_c \\ 
\end{bmatrix}.
$$
We then crop and resize the ROI using ROIAlign operator \cite{He_2017_ICCV}.
The whole process only involves matrix multiplication and pixel interpolation,
which is fully differentiable.
For the palm vein image,
we first translate it by $(\hat{d}_x,\hat{d}_y)$ to align the hands.
Next,
the ROI extraction of the palm vein image is straightforward.

\subsection{Bimodal Fusion Network}
The whole FNet is composed of several ResBlocks \cite{he2016deep} inserting a cross-modal selection module.
Before the fusion process,
we construct two light branches for preprocessing.
In the palmprint branch,
one ResBlock is used to convert $F$ into fusion space.
As the input of the palm vein branch is the ROI image,
for the sake of balance,
more blocks are added in the palm vein branch for feature extraction and downsampling.
The two branches joint at our proposed cross-modal selection module,
where the features have the same spatial dimension $(H_f, W_f)$.
After feature enhancement, 
two ResBlocks are added for further high-level feature extraction.

\textbf{Cross-Modal Selection.}
The vascular distribution endows the palm vein image the ability of anti-counterfeit,
while the palmprint image of one individual has more texture information and therefore is more distinctive.
In this work we focus more on person identification and treat the bimodal images differently in the fusion process.
Basically, the feature fusion should be guided by the palmprint feature.
To accord the idea,
we propose a selection scheme based on the channel correlation.
Suppose $P \in \mathbb{R}^{C_1\times H_f\times W_f}$,
$V \in \mathbb{R}^{C_2\times H_f\times W_f}$
are the palmprint feature and the palm vein feature respectively.
The correlation $r \in \mathbb{R}^{C_1\times C_2}$ between channels is defined as their cosine similariy:

\begin{equation}
\begin{aligned}
	r_{ij} = \frac{<P_{i}|V_{j}>}{||P_{i}||_2 \cdot ||V_{j}||_2}
\end{aligned}
\end{equation}
where the subscripts $1 \leq i \leq C_1$, $1 \leq j \leq C_2$ are the channel numbers and $<\cdot|\cdot>$ denotes the inner product.

For each channel of palmprint feature $P_i$,
it will select the $k$-$th$ palm vein feature $V_k$ which has maximum correlation to enhence its representation,
\ie $k = \max_{j} r_{ij}$.
Denote the selected feature as $V_i^s (V_i^s=V_k)$,
the fusion is the summation of two features:
\begin{equation}
\begin{aligned}
	P_i^f = P_i + V_i^s
\end{aligned}
\end{equation}
The channel-wise selection and summation enable the palm vein branch to learn the supplementary information for palmprint feature.
It should be pointed out that
if we conduct dynamic summation $\alpha P_i + \beta V_i^s$ with two learnable scalar parameters,
$\beta$ tends to 0 as the training goes on, 
which crashes the fusion.
After feature selection, 
the fusion network generates the final palmprint descriptor $E \in \mathbb{R}^{512}$.

\subsection{Network Training}

Following \cite{li20213d,zhang2020towards}, 
we adopt arc-margin loss \cite{deng2019arcface} on the top of the FNet to supervise the embedding $E$:
\begin{equation}
	\begin{aligned}
		L_{arc} =& -\log \frac{e^{s\cos(\eta_{y}+m)}}{e^{s\cos(\eta_{y}+m)}+\sum_{j=1,j\neq y}^ne^{s\cos(\eta_{j}+m)}} \\
	\end{aligned}
	\label{eq_arc}
\end{equation}
where $\eta$ is the angle between logit and weight in classification layer, 
$y$ denotes the ground truth label and $n$ is the number of classes.
In Eq. \eqref{eq_arc},
$s$ and $m$ are hyperparameters which represent scale and angular margin respectively.
The whole network is supervised by the following loss:
\begin{equation}
\begin{aligned}
	L_{total} = L_c + \lambda_1 L_r + \lambda_2(L_s+L_d) + \mu L_{arc}
\end{aligned}
\end{equation}
where $\mu, \lambda_1, \lambda_2$ are trade-off loss weights.
We set
$\mu=1,\lambda_1=0.1, \lambda_2=0.1$
in our experiments unless specified otherwise.

Since we use the predicted ROIs as input to the downstream, 
the training of FNet is misleading at the beginning (the palm is not well detected).
To avoid destroying FNet in the first few epochs,
we adopt a two-stage training strategy for BPFNet.
In stage I,
only DNet is trained, \ie $\mu=0$.
In stage II,
we optimize all the losses in BPFNet, \ie $L_{total}$, jointly.

\section{Experiments}
\label{sec_experiment}
In this section, 
we conduct experiments on CUHKSZ dataset and evaluate our method in various aspects.
We first exhibit the detection performance gap of two ROI extraction schemes.
Then we compare the performances of our BPFNet with other state-of-the-art methods,
where the recognition performances with or without palm vein image are reported.
Finally, we conduct an ablation study to analyze our fusion mechanism.

\subsection{Datasets and Metrics}
To evaluate our proposed method,
we conduct experiments on two touchless palmprint benchmarks CUHKSZ-v1 \cite{li20213d} and TongJi.
CUHKSZ-v1 has 28,008 palm images from 1167 individuals and TongJi has 12,000 images from 300 individuals.   
As for the dataset split, we follow the official train/test split and each palm is regarded as one class, 

We report IoU (Intersection over Union) in the ROI localization task. 
Rank-1 accuracy and EER (Equal Error Rate) are used for evaluating palmprint verification performance. 
In the evaluation,
we adopt a test-to-register protocol that considers the real application of palmprint verification.
The protocol is widely used in previous works \cite{matkowski2019palmprint,Genovese2019palmnet}.
Under this protocol,
four images are registered as enrollment and the remaining test images are matched to these images,
where the minimum distance of four distances is used as matching distance.

For the learning based methods,
the model is trained on the training set and evaluated on the test set.
Other methods are evaluated only on the test set for a fair comparison.

\subsection{Implementation} 
Our experiments are carried out on a server with 4 Nvidia TITAN RTX GPUs, 
an Intel Xeon CPU and 256G RAM.  
The proposed method is implemented by PyTorch \cite{pytorch2019} and code is available\footnote{\url{https://github.com/dxbdxx/BPFNet}}.
The DNet (based on ResNet18) is pretrained on ImageNet \cite{deng2009imagenet}
and all the layers in FNet are initialized by a Gaussian distribution of mean value 0 and standard deviation 0.01.
The final output stride of the detection network is 16.
We use the stochastic gradient descent (SGD) algorithm with momentum 5e-4 to optimize the total loss. 
The batch size for the mini-batch is set to 64. 
The initial learning rate for the CNN is 1e-2, 
which is annealed down to 1e-4 following a cosine schedule without restart \cite{loshchilov2017}. 
Stage I take 15 epochs in our experiment.
The total training epochs are 100.
No data augmentation is leveraged in our experiments.

\subsection{Detection Results} 
In this section,
we mainly discuss the effectiveness of DNet.
To demonstrate our detection scheme has better performance than keypoint detection based scheme,
we realize a keypoint detection method based on our network structure as comparison.
Note that the Center Head is in fact a keypoint detector,
we just need to change the number of keypoints from 1 to 4 (CUHKSZ has 4 keypoints).
Concretely,
we apply the same backbone network and only conserve the Center Head whose output channels are changed to 4.
The task of the detection network now is transferred to 4 keypoints detection 
and the ground truth heat map is also changed to 4 gaussian distributions,
where the maximum value will be set if they have overlap pixel.
The ROI is hence determined by the induced local coordinate system. 
We run experiments and the performances are shown in Table \ref{tab_iou},
which supports our claim.

\begin{table}[t]
    \begin{center}
        \scalebox{1}[1]{
        \begin{tabular}{|l||c|c|}
        \hline
        Scheme 						& CUHKSZ	& TongJi 	\\
        \hline
        Keypoint detection			& 84.32\%	& 87.68\%	\\
        Bounding box regression 	& 90.65\%	& 91.32\%	\\
        \hline
        \end{tabular}}
    \end{center}
	\caption{IoU comparison of Keypoint detection and Bounding box regression.}
    \label{tab_iou}
\end{table}

\begin{table}[t]
	\begin{center}
		\begin{tabular}{lccccc}
		\hline 
		\multirow{2}{*}{Methods} &\multirow{2}{*}{Test ROI}& \multicolumn{2}{c}{CUHKSZ}  & \multicolumn{2}{c}{TongJi} \\
		\cmidrule(lr){3-4}  \cmidrule(lr){5-6} 
						&				& Rank-1	& EER 	& Rank-1	& EER	\\
		\hline
		RGB 			& predicted		& 99.89 	& 0.15 	& 99.93  	& 0.22 	\\
		RGB + IR		& predicted		& \textbf{100} 	& \textbf{0.11} & \textbf{100} 	& \textbf{0.03}	\\
		\hline
		RGB 			& ground truth	& 99.47 	& 0.18 	& 99.72 	& 0.39 	\\
		RGB + IR		& ground truth	& 99.68 	& 0.14 	& 99.97 	& 0.05	\\
		\hline
		\end{tabular}
	\end{center}
	\caption{The performance comparison (\%) of single RGB image input versus bimodal image input.}
\label{tab_input}
\end{table}

\subsection{Bimodal fusion} 
The bimodal fusion leverages RGB image and IR image simultaneously,
where the IR image provides supplementary information for person identification.
For showing the effect of the bimodal fusion scheme, 
as BPFNet is separable,
we conduct palmprint recognition experiments using single RGB images as comparison.
For eliminating the possible influence of the ROI bias,
we also evaluate our model with ground truth ROIs.
It should be noted that we do not acquire the ground truth ROIs in the training.
The experiment results are shown in Table \ref{tab_input}.
We can see that the palm vein image can further improve the recognition performance.

An interesting point is that the performance of our model with predicted ROI is better than that with ground truth ROI.
There are two reasons for this phenomenon. 
First,
BPFNet is a unified framework,
the ROI localization is not perfect but has already achieved high accuracy (IoU more than 90\%),
which will not hinder further feature extraction. 
Second,
FNet is already adapted to the ROIs extracted by DNet, 
however the data distribution of ground truth ROIs and extracted ROIs are different.
Therefore FNet has better performance when input predicted ROIs.

\begin{table}[t]
	\centering
	\begin{tabular}{lcccc}
	\toprule 
	\multirow{2}{*}{Methods} & \multicolumn{2}{c}{GT ROI}  & \multicolumn{2}{c}{Predicted ROI} \\
	\cmidrule(lr){2-3}  \cmidrule(lr){4-5} 
						& Rank-1  	& EER 	& Rank-1   	& EER	\\
	\toprule
	CompCode 			& 99.83		& 0.32 	& 96.67  	& 2.88 	\\
	OrdinalCode			& 99.79 	& 0.42 	& 95.23 	& 3.39	\\
	LLDP$_3$	   		& 99.68 	& 0.44 	& 95.12 	& 2.64 	\\
	CR-CompCode 		& 99.79		& 0.32 	& 97.67	   	& 2.06 	\\
	\hline
	Resnet18 			& 99.25		& 0.58 	& 98.41 	& 0.66 	\\
	VGG11-bn 			& 90.89 	& 2.50 	& 95.55 	& 1.37	\\
	GoogLeNet 			& 84.11 	& 2.97 	& 80.19 	& 2.81	\\
	PalmNet 	   		& 95.97 	& 0.79 	& 96.21 	& 1.48	\\
	BPFNet$_{rgb}$		& 99.47 	& 0.18	& \textbf{99.89} & \textbf{0.15}  \\
	\bottomrule
	\end{tabular}
	\caption{The performance comparison (\%) on CUHKSZ dataset with ground truth ROI (GT ROI) 
			or ROI extracted by BPFNet (Predicted ROI).}
	\label{tab_RGB}
\end{table}

\subsection{Comparison with other Methods}
\begin{figure*}[t]
	\centering
	\includegraphics[width=\linewidth]{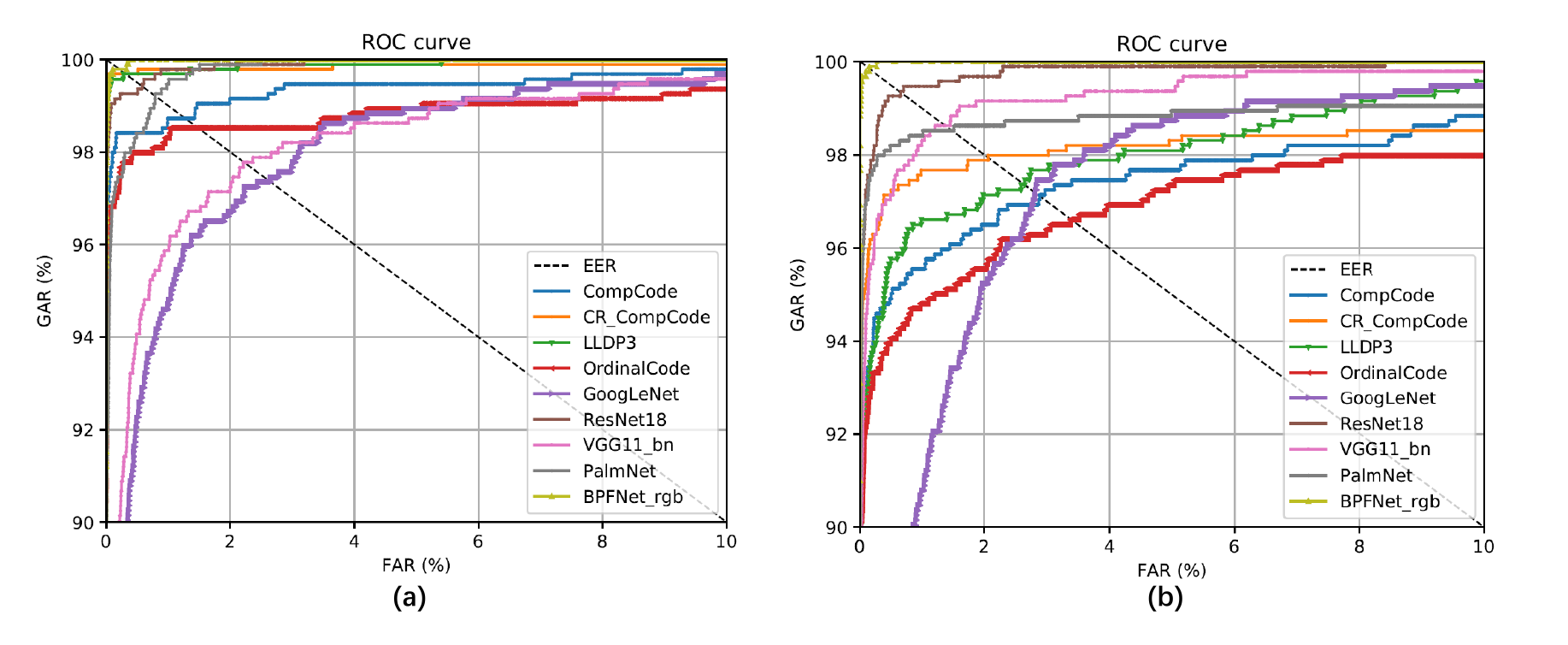}
	\caption{The ROC curves obtained using the methods listed in Table \ref{tab_RGB}. 
			The figure shows the results of methods
			(a) with ground truth ROI as input and
			(b) with our extracted ROI as input.  
			}
\label{fig_roc} 
\end{figure*}

Since CUHKSZ dataset is more challenging and the comparison results are more clear,
after we will conduct experiments only on CUHKSZ dataset.
We compare our method with coding based methods including CompCode \cite{Kong2004Competitive}, OrdinalCode \cite{sun2005ordinal}, LLDP \cite{luo2016local}, CR-CompCode \cite{zhang2017towards} 
as well as deep learning method PalmNet \cite{Genovese2019palmnet} and several baselines \cite{he2016deep,simonyan2014very,szegedy2015going}.
All deep learning baselines are trained with arc-margin loss.
These methods use palmprint ROI images as input and do not involve palm vein image fusion. 
For a fair comparison,
we report the performances of our model trained with only RGB images, 
denoted as BPFNet$_{rgb}$.
The IR image is not used during training/test and hence the cross-modal selection scheme does not contribute.
Moreover, 
considering the possible influence of ROI bias,
we compare both the performances using ROI extracted by BPFNet and the performances using ground truth ROI,
as shown in Table \ref{tab_RGB}.
Some ground truth ROIs and predicted ROIs are shown in Supplementary Materials for visualization.
The corresponding ROC curves are plotted in Fig. \ref{fig_roc}.

\subsection{Ablation Study}
To demonstrate the effectiveness of our cross-modal selection module,
we set several baseline fusion methods as comparison.
The results are shown in Table \ref{tab_fusion}.
In the table,
``Average" and ``Max" means the fusion is conducted with element-wise average and element-wise maximum operation respectively.
We can see that our fusion strategy achieves the best performances.

\begin{table}
\begin{center}
	\scalebox{1}[1]{
	\begin{tabular}{lccc}
	\toprule
	Strategy 				&& Rank-1 	& EER	\\
	\toprule
	RGB 					&& 99.89 	& 0.15  \\
	RGB+IR Average 		  	&& 99.50 	& 0.23  \\
	RGB+IR Max 				&& 99.15 	& 0.21 	\\
	cross-modal selection   && 100    	& 0.11	\\
	\bottomrule
	\end{tabular}}
\end{center}
\caption{The performance comparison (\%) of different fusion strategies.}
\label{tab_fusion}
\end{table}

\section{Conclusion}
\label{sec_conclusion}
In this paper, 
we propose a novel framework, named BPFNet,
for bimodal palmprint recognition.
In our method,
the detection network directly regresses the rotated bounding box,
which makes it compatible with other annotation systems.
In the downstream,
the fusion network conducts feature fusion using the proposed cross-modal selection.
Finally, comprehensive experiments are carried out to demonstrate the superiority of our method.

\bibliography{mybibfile}

\end{document}